# Extreme learning machine-based model for Solubility estimation of hydrocarbon gases in electrolyte solutions

**Narjes Nabipour** [1], **Amir Mosavi** [2,3,4,5], **Alireza Baghban** [6], **Shahaboddin Shamshirband** [7,8*], **Imre Felde** [9]

[1] Institute of Research and Development, Duy Tan University, Da Nang 550000, Vietnam; narjesnabipour@duytan.edu.vn

[2] Kando Kalman Faculty of Electrical Engineering, Obuda University, 1034 Budapest, Hungary; amir.mosavi@kvk.uni-obuda.hu

[3] School of Built the Environment, Oxford Brookes University, Oxford OX30BP, UK

[4] Queensland University of Technology, Faculty of Health, 130 Victoria Park Road, Queensland 4059, Australia; a.mosavi@qut.edu.au

[5] Institute of Structural Mechanics, Bauhaus Universität-Weimar, D–99423 Weimar, Germany; amir.mosavi@uni-weimar.de

[6] Chemical Engineering Department, Amirkabir University of Technology, Mahshahr Campus, Mahshahr, Iran

[7] Department for Management of Science and Technology Development, Ton Duc Thang University, Ho Chi Minh City, Viet Nam; shahaboddin.shamshirband@tdtu.edu.vn

[8] Faculty of Information Technology, Ton Duc Thang University, Ho Chi Minh City, Viet Nam

[9] John von Neumann Faculty of Informatics, Obuda University, 1034 Budapest, Hungary

**Abstract-** Calculating hydrocarbon components solubility of natural gases is known as one of the important issues for operational works in petroleum and chemical engineering. In this work, a novel solubility estimation tool has been proposed for hydrocarbon gases including methane, ethane, propane and butane in aqueous electrolyte solutions based on extreme learning machine (ELM) algorithm. Comparing the ELM outputs with a comprehensive real databank which has 1175 solubility points concluded to R-squared values of 0.985 and 0.987 for training and testing phases respectively. Furthermore, the visual comparison of estimated and actual hydrocarbon solubility leaded to confirm the ability of proposed solubility model. Additionally, sensitivity analysis has been employed on the input variables of model to identify their impacts on hydrocarbon solubility. Such a comprehensive and reliable study can help engineers and scientists to successfully determine the important thermodynamic properties which are key factors in optimizing and designing different industrial units such as refineries and petrochemical plants.

Keywords: hydrocarbon gases; solubility; extreme learning machines; deep learning; electrolyte solution; prediction model

## 1. Introduction

Solubility of hydrocarbon and nonhydrocarbon gases, i.e., mixtures of methane, ethane, propane, $CO_2$ and $N_2$ in aqueous phases, is known as one of the important practical and theoretical challenges in petroleum, geochemical and chemical engineering. This property has effective role in different processes such as achieving optimum conditions for oil and gas transportation, gas hydrate formation, designing thermal separation processes, gas sequestration for protecting environment, and coal gasification. Petroleum reservoirs normally have some natural gases with aqueous solution at high-pressure and high-temperature conditions so that the solubility of gas becomes attractive for engineers [1-8]. In production and transportation of hydrocarbons, it is possible that water content of gas undergoes an alteration in phase from vapor to ice and gas hydrates. The crystalline solid phases called gas hydrates are created when small sized gas molecules are trapped in lattice of water molecules. Creation of hydrates can cause major flow assurance problems during production and transportation of hydrocarbons steps such as pipeline blockage, corrosion and many other issues resulted from the two-phase flow [1, 9-11].

In the recent years, investigations on $CO_2$ solubility in aqueous electrolyte solutions have grown significantly as well as they are related to $CO_2$ capture and storage. It is a clear fact that the dominant cause of global warming is emission of $CO_2$ gas generated from fossil fuels so its sequestration and disposal in the ocean have been known as a reasonable choice to overcome global warming problems [12-14]. Simulation of enhanced oil recovery, design of supercritical extraction and optimization of $CO_2$ dissolution in the ocean need a comprehensive knowledge about carbon dioxide solubility in aqueous electrolytes solutions [13-15].

Investigation of natural gas phase behavior in aqueous solutions in different operational conditions is known one of the important issues in the industry, which has wide applications for avoiding problems in designing and optimization of gas processing. In the literature, there are different solubility datasets for various gas-liquid systems. These datasets mostly include hydrocarbons' dissolution in water/brine systems [1, 4, 5, 9, 16-20] and non-hydrocarbons such as $CO_2$ and $N_2$ dissolution in water/brine systems [7, 12-14, 18, 21-24]. A brief summary of the hydrocarbon systems datasets is shown in **Table 1** for hydrocarbons. The experimental data of water content of hydrocarbons and non-hydrocarbons are limited because of difficulties in measurement of the low water content gases at high pressure and low temperature. Mohammadi and coworkers expressed that an accurate estimation of water content can be obtained by gas solubility data, therefore, they overcame the complexities of experimental determination of the water content in natural gases [1]. Due to limited number of measurement data, wide attempts have been made to model and describe the gas-liquid equilibrium in aqueous electrolyte solutions. There are several thermodynamic models which uses the Henry's constant, activity coefficient and cubic equations of state to obtain more information about the equilibrium conditions. The changes of Henry's constant for the pressure lower than 5 MPa are negligible and it is dominantly effected by temperatures[19]. The high dependency on temperature is obvious at low temperature and also the nonlinear decreasing relationship is observed at high temperatures [25]. Furthermore, there is just a limited number of Henry's constants for hydrocarbon systems at low temperature. According to this fact, there are several drawbacks in applying the Henry's law, whereas it has great ability in accurate prediction of solubility. As an example, it is suitable for dilute solutions or near ideal solutions [26]. Additionally, this method is correct for single compounds in no chemical reaction conditions for aqueous phase. Another method is cubic EOS which has several advantages such as small number of parameters, computational efficiency and ease of performance [3, 4, 21]. The EOSs were proposed originally for pure fluids, after that, their applications were expanded for mixtures by combining the constants from different pure components. This extension can be done by different methods such as Dalton's law of additive partial pressures and Amagat's rule of additive volumes[5]. For complex compounds, there are some limitations in accuracy of EOS which highlight the importance of empirical adjustments by dealing with the binary interaction parameters. In order to determine these parameters, a reliable source of experimental data for vapor-liquid equilibrium is required which induces some uncertainty into EOSs[7].

Due to above discussions, development of an accurate and reliable approach for estimation of solubility of hydrocarbons and non-hydrocarbons in aqueous electrolyte solutions has been highlighted. Nowadays, machine learning approaches have shown extensive applications in different topics [27-35]. This work organizes a novel artificial intelligence method called Extreme Learning Machine (ELM) to estimate solubility of hydrocarbons in aqueous electrolyte mixtures in terms of types of gas, mole fractions of gases, pressure, temperature and ionic strength.

**Table 1.** Details of experimental hydrocarbons solubility in aqueous electrolyte solutions.

| Author | P(Mpa) | T(°C) | composition | Mole fraction of the components in the gaseous phase |
|---|---|---|---|---|
| Culberson et al. | 0.8-69.61 | 37.78-171.11 | Pure water | C1: 0.0000698-0.0033 |
| Kiepe et al. | 0.304-10.23 | 40-100.14 | Pure water, LiBr, KBr, LiCl, KCl | C1:0.00003-0.00154 |
| Chapoy et al. | 0.357-18 | 1.98-95.01 | Pure water | C1:0.000204-0.002459 |
| | | | | C2:0.0000147-0.0000674 |
| | | | | C3:0.0000321-0.0002694 |
| | | | | C4:0.00000387-0.00001121 |
| Marinakis et al. | 6.22-20.1 | 1.4-25.98 | Pure water, NaCl | C1:0.00099-0.00282 |
| | | | | C2:0.000038-0.000249 |
| | | | | C3:0.000006-0.000042 |
| Crovetto et al. | 1.327-6.451 | 24.35-245.15 | Pure water | C1: 0.0002124–0.0010337 |
| Wang et al. | 1-40.03 | 2.5-30.05 | Pure water | C1: 0.000563–0.004049 |
| | | | | C2: 0.0000986–0.000864 |

| | | | | C1: 0.00045–0.0037 |
|---|---|---|---|---|
| Amirjafari | 4.66-56.16 | 54.44-104.44 | Pure water | C2: 0.000119–0.001768 |
| | | | | C3: 1.9e−5–0.001863 |
| O'Sullivan et al. | 10.2-62 | 51.5-125 | Pure water, NaCl | C1: 0.000805–0.0043 |
| | | | | C2: 0.000825–0.001438 |
| Michels et al. | 4.09-45.89 | 25-150 | Pure water, NaCl, LiCl, NaBr, NaI, CaCl$_2$ | C1: 0.000173–0.00269 |
| Mohammadi et al. | 1.14-31.1 | 4.65-24.75 | Pure water | C1: 0.000313–0.00311 |
| Vul'fson et al. | 2.53-60.8 | 19.95-79.95 | Pure water | C1: 0.000361–0.004328 |
| Dhima | 2.5-100 | 71 | Pure water | C1: 0.000127–0.005085 |
| | | | | C2: 0.000821–0.001398 |
| | | | | C4: 0.000021–0.000103 |

## 2. Methodology

### 2.1. Experimental dataset collection

In order to construct a highly accurate and comprehensive model capable of estimating the solubility of mixtures of hydrocarbons in aqueous electrolyte solutions, a comprehensive databank was provided based on existing experimental data in **Table 1**. This databank contains total number of 1175 solubility points for hydrocarbons (881 and 294 points for training and testing phases respectively.). According to the literature [1, 4, 5, 9, 16-20], the solubility of gases in these systems is highly function of aqueous solutions, pressure, temperature and gaseous phase composition. The aqueous phase composition was change into ionic strength (I) from salt concentrations to reduce dimensions of modeling process. The following equation presents the relationships between ionic strength, valance of charged ions ($z_i$) and molar concentration of each ion ($m_i$):

$$I = \frac{1}{2}\sum m_i |z_i|^2 \quad (1)$$

In this study, the solubility of hydrocarbons is predicted in terms of concentration of components in gaseous mixture, ionic strength of solution, temperature and pressure:

$$\eta_h = f(C1, C2, C3, C4, I, P, T, idx) \quad (2)$$

In which, $\eta_h$ represents the hydrocarbon solubility in aqueous phase; C(1-4) are known as the methane, ethane, propane and butane mole fraction in gas phase(0-99.99); I denotes the ionic strength based on molarity(0-37.35); T denotes the temperature in terms of °C (1.4-245.15); P shows the pressure in MPa (0.3-100), and idx symbolizes the index of fraction whose solubility is to be determined (1,2,3,4).

*2.2. Extreme Learning Machine*

Huang proposed a new intelligence method based on single-layer feedforward neural network (SLFFNN) called Extreme Learning Machine to satisfy the drawbacks of gradient-based algorithms such low training speed and low learning rate. In the ELM algorithm, the hidden nodes are selected randomly and the weights of output of the SLFFNN are calculated by applying Moore-Penrose generalized inverse [36, 37].

The scheme of ELM algorithm is demonstrated in **Figure 1**. By assuming N training sets such as $(x_i, y_i) \in R^n \times R^m$ for L hidden nodes, the SLFFNN algorithms can be written as following:

$$\sum_{i=1}^{L} \beta_i f_i(x_j) = \sum_{i=1}^{L} \beta_i f_i(a_i . b_i . x) \qquad j=1,\ldots,N \quad (3)$$

In which, $a_i = [a_{i1},\ldots,a_{in}]^T$ points to input weights matrix which is related to hidden nodes, $\beta_i = [\beta_{i1},\ldots,\beta_{im}]^T$ represents the output weights matrix which is related to hidden nodes, and $b_i$ symbolizes the hidden layer bias.

$$\sum_{i=1}^{L} \beta_i f_i(x_j) = H\beta \qquad (4)$$

In which, $\beta = [\beta_1,\ldots,\beta_L]$ and $h(x) = [h_1(x),\ldots h_L(x)]$ are known as the hidden layer output matrix and the output weight matrix.

The first step of this model is the random calculation of input weight and the bias of hidden layer for the training phase. Then, determination these values, the hidden layer matrix is obtained by utilization of input variables. Then, the SLFFNN training is changed to a least square problem. The ELM algorithms implement regularization theory to define a target function as following [38-40]:

$$\min L_{ELM} = \frac{1}{2}||\beta||^2 + \frac{c}{2}||T - H\beta||^2 \qquad (5)$$

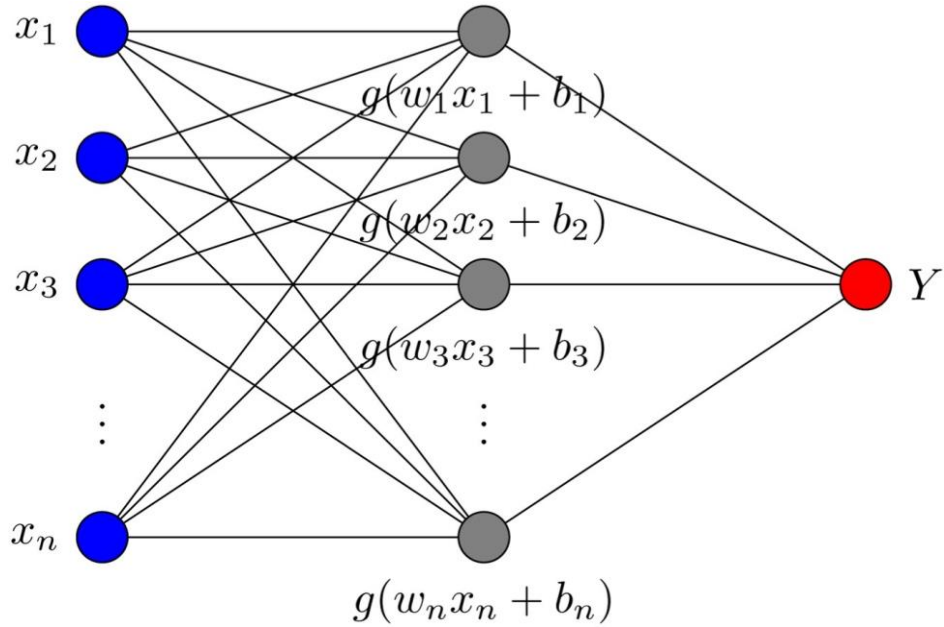

**Figure 1.** Structure of ELM algorithm.

## 3. Results and discussion

In this study, the solubility of hydrocarbons in the aqueous electrolyte phase is determined based on ELM algorithm. To this end, the sigmoid function is set as activation function and the input weights were initialized randomly in range of (-1,1). Additionally, the number of nodes in the hidden layers was estimated as 30 based on the lowest value of RMSE as determined in **Figure 2**. As shown, after 30 nodes, by increasing complexity of model, the testing error increased so the optimum structure of the algorithm has 30 nodes to prohibit overfitting.

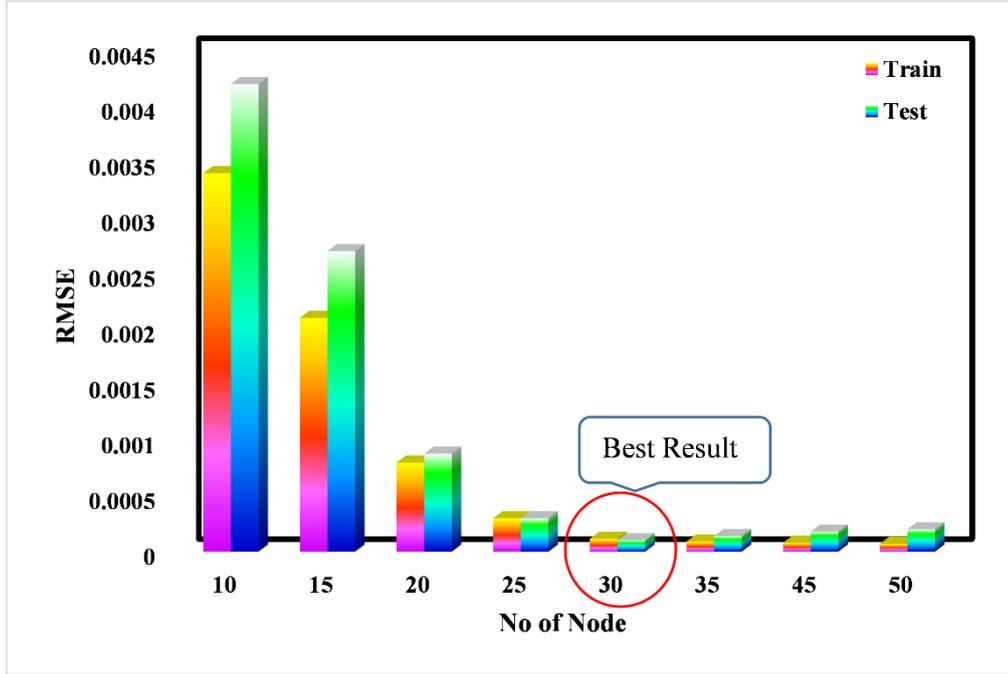

**Figure 2.** Obtaining optimum structure of proposed algorithm

In the following, the statistical results of the estimation of hydrocarbon solubility are inserted in **Table 2**. The below equations are used to achieve this end:

$$\text{Mean relative error (MRE)} = \frac{100}{N} \sum_{i=1}^{N} \left( \frac{X_i^{actual} - X_i^{predicted}}{X_i^{actual}} \right) \tag{6}$$

$$\text{Root mean square error (RMSE)} = \sqrt{\frac{1}{N} \sum_{i=1}^{N} \left( (X_i^{actual} - X_i^{predicted})^2 \right)} \tag{7}$$

$$\text{Mean squared error (MSE)} = \frac{1}{N} \sum_{i=1}^{N} (X_i^{actual} - X_i^{predicted})^2 \tag{8}$$

$$\text{R-squared } (R^2) = 1 - \frac{\sum_{i=1}^{N}(X_i^{actual} - X_i^{predicted})^2}{\sum_{i=1}^{N}(X_i^{actual} - X^{actual})^2} \tag{9}$$

As shown in **Table 2**, the MRE, MSE and RMSE are determined as 22.049, 1.33285E-08 and 0.0001 for training phase respectively. Moreover, for testing phase, MRE=22.054, MSE=1.05351E-08 and RMSE=0.0001 are calculated. The estimated $R^2$ values are 0.985, 0.987 and 0.985 for training, testing and overall datasets respectively. These results give the knowledge about the high degree of accuracy for proposed ELM algorithm.

**Table 2.** The statistical analyses of developed model.

| Dataset | $R^2$ | MRE (%) | MSE | RMSE |
|---|---|---|---|---|
| Training | 0.985 | 22.049 | 1.33285E-08 | 0.0001 |
| Testing | 0.987 | 22.054 | 1.05351E-08 | 0.0001 |
| Overall | 0.985 | 22.050 | 1.26295E-08 | 0.0001 |

On the one hand, the comparison between the estimated and real hydrocarbons solubility in aqueous electrolyte solutions are shown in **Figure 3**. This depiction demonstrates an excellent agreement between estimated and real solubility values. **Figure 4** also represents the regression plot of actual hydrocarbons solubility versus estimated one. A light cloud of data near the 45º line expresses the validity and accuracy of ELM algorithm. Additionally, **Figure 5** also shows the distribution of relative deviations between forecasted and actual hydrocarbons solubility in aqueous solutions. It can be seen that the ELM outputs deviate slightly from the real solubility and most of relative deviations are near to zero. Furthermore, **Figure 6** shows the histograms of relative deviations for training and testing phases. In this demonstration, frequency diagram confirms that most of the error points are close to zero and also cumulative axis express the fact that range of deviation is very limited and the highest slop of cumulative curve occurred near the zero point.

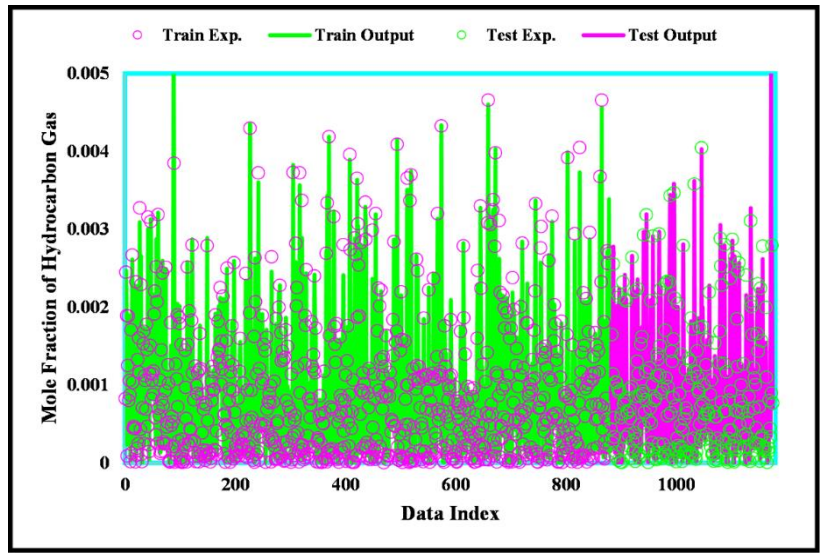

**Figure 3.** Comparison of actual and estimated solubility of hydrocarbons.

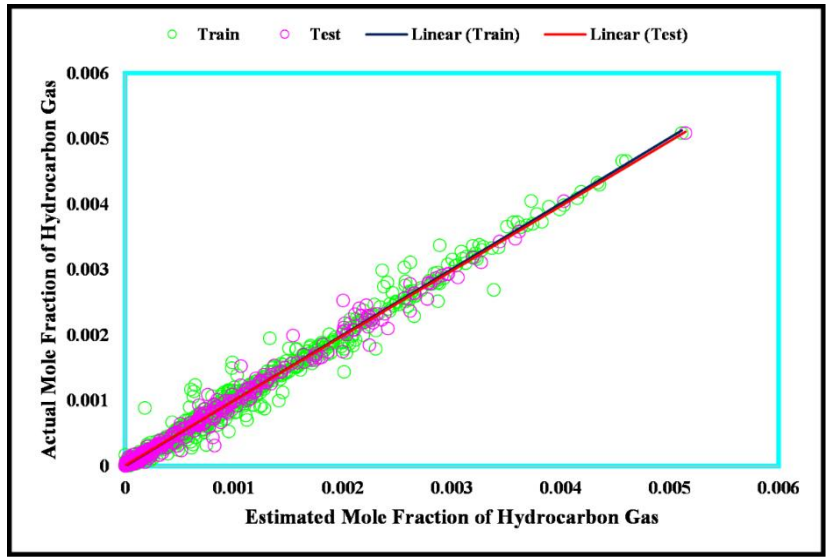

**Figure 4.** Cross plot of actual and estimated solubility of hydrocarbons.

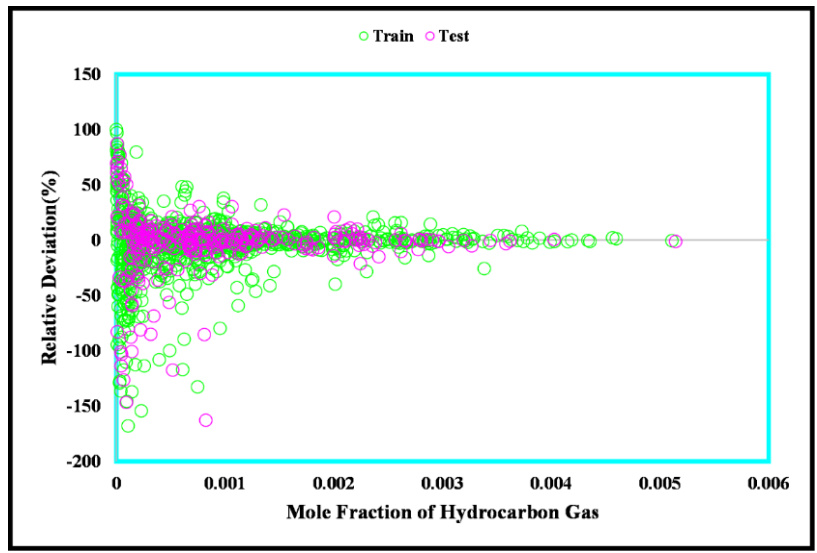

**Figure 5.** Relative deviation between actual and estimated solubility of hydrocarbons.

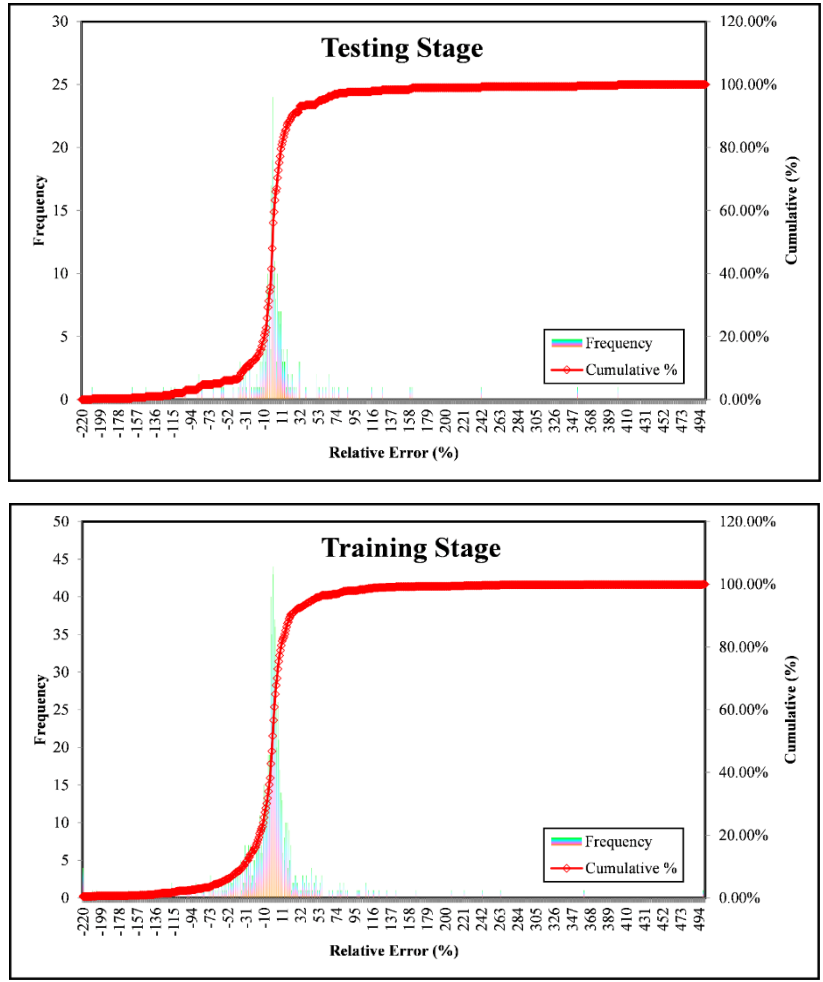

**Figure 6.** Histogram diagram of relative deviations.

The ELM algorithm implemented in the current work shows an excellent ability in calculation of solubility of hydrocarbons in aqueous phases. One of the important factors which can influence the validation of model is degree of precision of utilized data. In order to clarify the accuracy of solubility databank, the Leverage mathematical method is recruited. This method has some rules to identify the suspected solubility data so that a matrix which is known as Hat matrix, should be constructed based on following formulation[41-45]:

$$H = U(U^T U)^{-1} U^T \qquad (10)$$

In which, U symbolizes a matrix of i*j dimensional. i and j are known as the number of algorithm parameter and training points which are used for determination of critical leverage limit as following:

$$H^* = 3(j+1)/i \qquad (11)$$

In order to detect the reliable zone, there are two standard residual indexes (-3 and 3) which are used in the leverage method. As shown in **Figure 7**, the reliable area is bound by these two residual indexes and critical leverage limit. The critical straight lines are shown by red and green colors. This plot is known as William's plot. In this plot, normalized residual is depicted versus hat value which is determined from the main diagonal of aforementioned matrix. It is obvious that the major number of solubility data are located in this area which expresses validation of hydrocarbon solubility databank.

In the most of parametric studies, it is a valuable attempt to identify the effectiveness of all inputs on the target. According to this fact, the sensitivity analysis is employed to investigate effect of concentration of components in gaseous mixture, ionic strength of solution, temperature and pressure on solubility of hydrocarbons in aqueous electrolyte systems. To this end, relevancy factor should be determined as following for each input parameter [46-52]:

$$r = \frac{\sum_{i=1}^{n}(X_{k,i}-\overline{X_k})(Y_i-\overline{Y})}{\sqrt{\sum_{i=1}^{n}(X_{k,i}-\overline{X_k})^2 \sum_{i=1}^{n}(Y_i-\overline{Y})^2}} \quad (12)$$

In which $Y_i$ and $\overline{Y}$ denote the 'i' th output and output average. $X_{k,i}$ and $\overline{X_k}$ are known as 'k'th of input and average of input. **Figure 8** shows the relevancy factor for each effective variable of hydrocarbon solubility. It is necessary to explain that the relevancy factor lies in range of -1 to1 so that the higher absolute value has more impact on hydrocarbon solubility. Furthermore, the positive relevancy factor shows the straight relationship between input and target. The relevancy factors for pressure, temperature, the index of fraction, ionic strength, methane, ethane, propane and butane mole fraction in gas phase are 0.52, 0.20, -0.48, -0.16, 0.11, 0.06, -0.19 ,and -0.07 respectively.  According to this explanation and results, as pressure, temperature, and mole fraction of methane and ethane increase, the solubility of investigated hydrocarbon increases. Moreover, pressure and mole fraction of ethane in gaseous phase are the most and least effective parameters on determination of solubility of hydrocarbons.

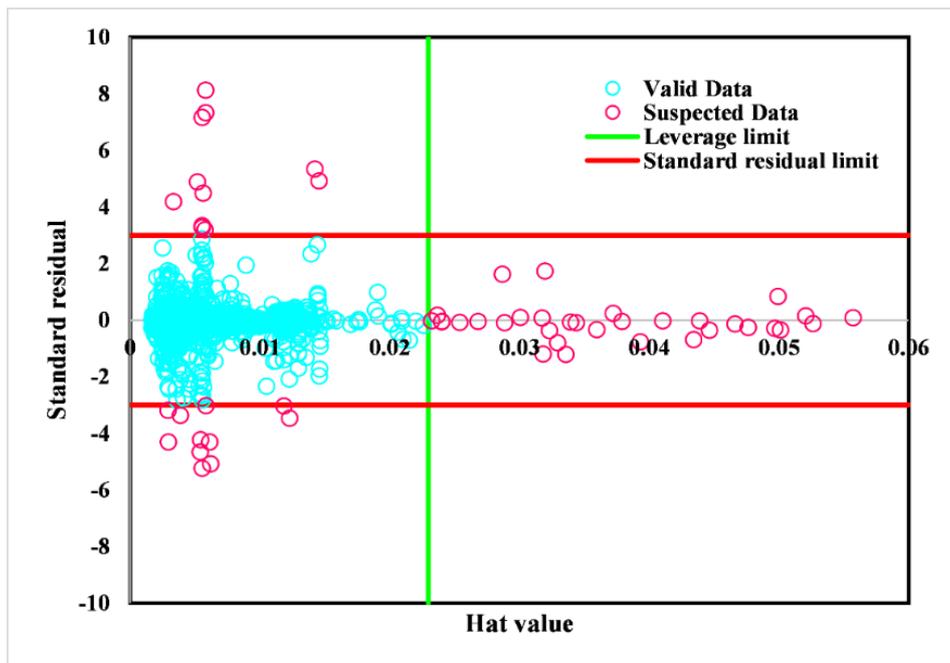

**Figure 7.** Detection of suspected data for hydrocarbons solubility dataset.

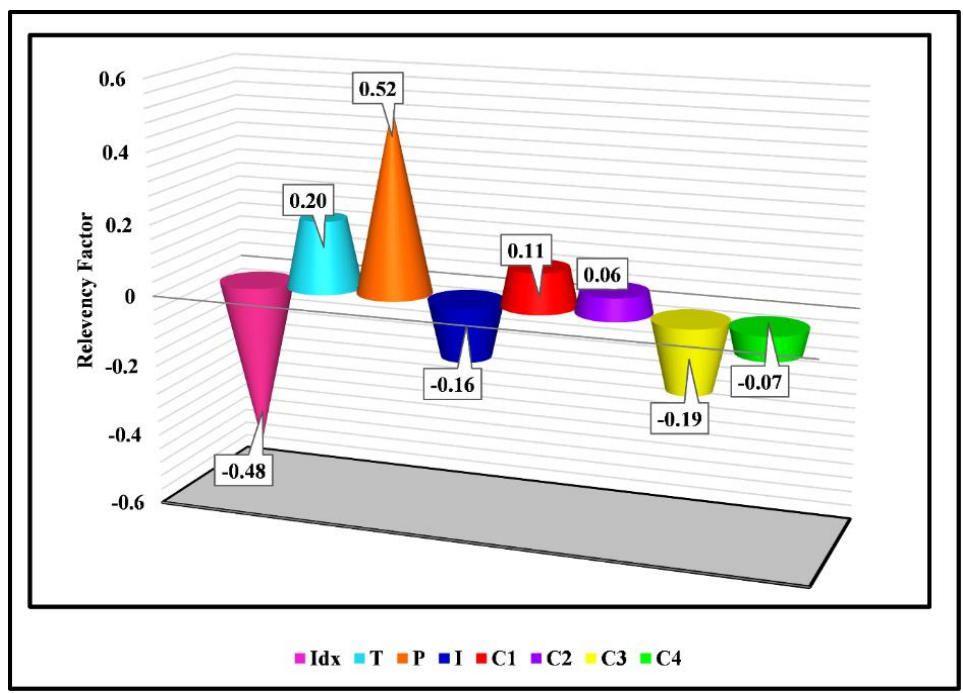

**Figure 8.** Sensitivity analysis for solubility of hydrocarbons.

## 4. Conclusion

The hydrocarbons solubility in aqueous electrolyte phases at high temperature and pressure conditions is known as a major effective parameter in variety of applications for petroleum industries and chemical engineering. Numerous attempts have been made in the current study to suggest a highly accurate and comprehensive predicting tool on the basis of Extreme Learning Machine to calculate hydrocarbons solubility in wide ranges of operational conditions. Comparing the ELM outputs with a comprehensive real databank which has 1175 solubility points concluded to R-squared values of 0.985 and 0.987 for training and testing phases respectively. The excellent agreements of ELM and real hydrocarbon solubility values express that the ELM algorithm is a valuable tool for designing and optimization of various processes that are relating to vapor-liquid equilibrium. Furthermore, this study gives more information about the intensity of each input parameter on solubility of hydrocarbons. Due to the aforementioned results, this work have potential to use in commercial software packages such as CMG and ECLIPSE for simulation of fluid flow in porous media.

**Acknowledgements:** we acknowledge the financial support of this work by the Hungarian State and the European Union under the EFOP-3.6.1-16-2016-00010 project.

**Author Contributions:** Conceptualization, A.B., and S.S.; methodology, N.N., A.M., and A.B.; software, N.N., A.M., A.B., S.S. and I.F.; validation, N.N., A.M., A.B., S.S. and I.F.; formal analysis, N.N., A.M., A.B., S.S. and I.F.; investigation, N.N., A.M., A.B., S.S. and I.F.; resources, N.N., A.M., A.B., S.S. and I.F.; data curation, N.N., A.M., A.B., S.S. and I.F.; writing—original draft preparation, N.N., A.M., and A.B.,; writing—review and editing, N.N., A.M., A.B., S.S. and I.F.; visualization, N.N., A.M., A.B., S.S. and I.F.; supervision, I.F.; project administration, A.M.

**Conflicts of Interest:** The authors declare no conflicts of interest.